%% file: Formatting-Instructions-LaTeX-2021.tex
\pdfoutput=1
\def\year{2021}\relax
\documentclass[letterpaper]{article} 
\usepackage{aaai21}  
\usepackage{times}  
\usepackage{helvet} 
\usepackage{courier}  
\usepackage[hyphens]{url}  
\usepackage{graphicx} 
\usepackage{natbib}  
\usepackage{caption} 
\usepackage{url}
\usepackage{multirow}
\usepackage{amsmath}

\usepackage{amsthm}
\usepackage{amssymb}
\usepackage{array}
\usepackage{booktabs}
\usepackage[table,xcdraw,dvipsnames]{xcolor}
\usepackage[ruled,linesnumbered, vlined]{algorithm2e}

\makeatletter
\newcommand{\removelatexerror}{\let\@latex@error\@gobble}
\makeatother

\author{
    Tongtong Wu\textsuperscript{\rm 1}\thanks{This work has been done by Tongtong Wu during the visiting period at Monash University, and the original idea was generated in the internship at Gamma Lab, Ping An OneConnect.}, 
    Xuekai Li\textsuperscript{\rm 1},
    Yuan-Fang Li\textsuperscript{\rm 2},
    Gholamreza Haffari\textsuperscript{\rm 2},
    Guilin Qi\textsuperscript{\rm 1}\thanks{Contact author.},\\
    Yujin Zhu\textsuperscript{\rm 3},
    Guoqiang Xu\textsuperscript{\rm 3}
    \\
}
\affiliations{
    \textsuperscript{\rm 1}School of Computer Science and Engineering, Southeast University, Nanjing, China\\
    \textsuperscript{\rm 2}Faculty of Information Technology, Monash University, Melbourne, Australia\\
    \textsuperscript{\rm 3}Gamma Lab, Ping An OneConnect, Shanghai, China\\
    \{wutong8023,lxk,gqi\}@seu.edu.cn, \{yuanfang.li,Gholamreza.Haffari\}@monash.edu,\\
    \{zhuyujin204, xuguoqiang371\}@ocft.com

}

\title{Curriculum-Meta Learning for Order-Robust Continual Relation Extraction}

\begin{document}

\maketitle

\begin{abstract}
Continual relation extraction is an important task that focuses on extracting new facts incrementally from unstructured text. 
Given the sequential arrival order of the relations, this task is prone to two serious challenges, namely catastrophic forgetting and order-sensitivity. We propose a novel curriculum-meta learning method to tackle the above two challenges in continual relation extraction. We combine meta learning and curriculum learning to quickly adapt model parameters to a new task and to reduce interference of previously seen tasks on the current task. 
We design a novel relation representation learning method through the distribution of domain and range types of relations. 
Such representations are utilized to quantify the difficulty of tasks for the construction of curricula. Moreover, we also present novel difficulty-based metrics to quantitatively measure the extent of order-sensitivity of a given model, suggesting new ways to evaluate model robustness. 
Our comprehensive experiments on three benchmark datasets show that our proposed method outperforms the state-of-the-art techniques. 
The code is available at the anonymous GitHub repository \url{https://github.com/wutong8023/AAAI-CML}.
\end{abstract}

\input{sec1-intro}
\input{sec2-related}

\input{sec4-framework}

\input{sec5-knowledge}

\input{sec6-expr}
\input{sec7-conc}

\section*{Acknowledgments}
Research in this paper was partially supported by the National Key Research and Development Program of China under grants (2018YFC0830200, 2017YFB1002801), the Natural Science Foundation of China grants (U1736204), the Judicial Big Data Research Centre, School of Law at Southeast University. 

\bibliography{bibfile}


\end{document}


\maketitle

\section*{A. Workflow}
our proposed curriculum-meta learning framework directed by knowledge-based curricula. During the learning phase, at each time step $t$, (1) meta-learner $L$ fetches the initialization parameters $\theta_t$ from the memory to initialize the model $f_{\theta_t}(.)$. (2) $L$ replays on the curriculum set $C_t$ which is sampled and sorted by the knowledge-based curriculum module KB-C. (3) $L$ trains on the support set $S_t$ of the current task $\mathcal{T}_t$. (4) Finally, $L$ updates the learned parameters $\theta_{t+1}$ and stores a small number of prototype instances of the current task into the memory. In the evaluation phase, the trained model is given a target set with labeled unseen instances from all observed tasks.

\begin{figure}[]
    \centering
    \resizebox{8.5cm}{!}{
    \includegraphics{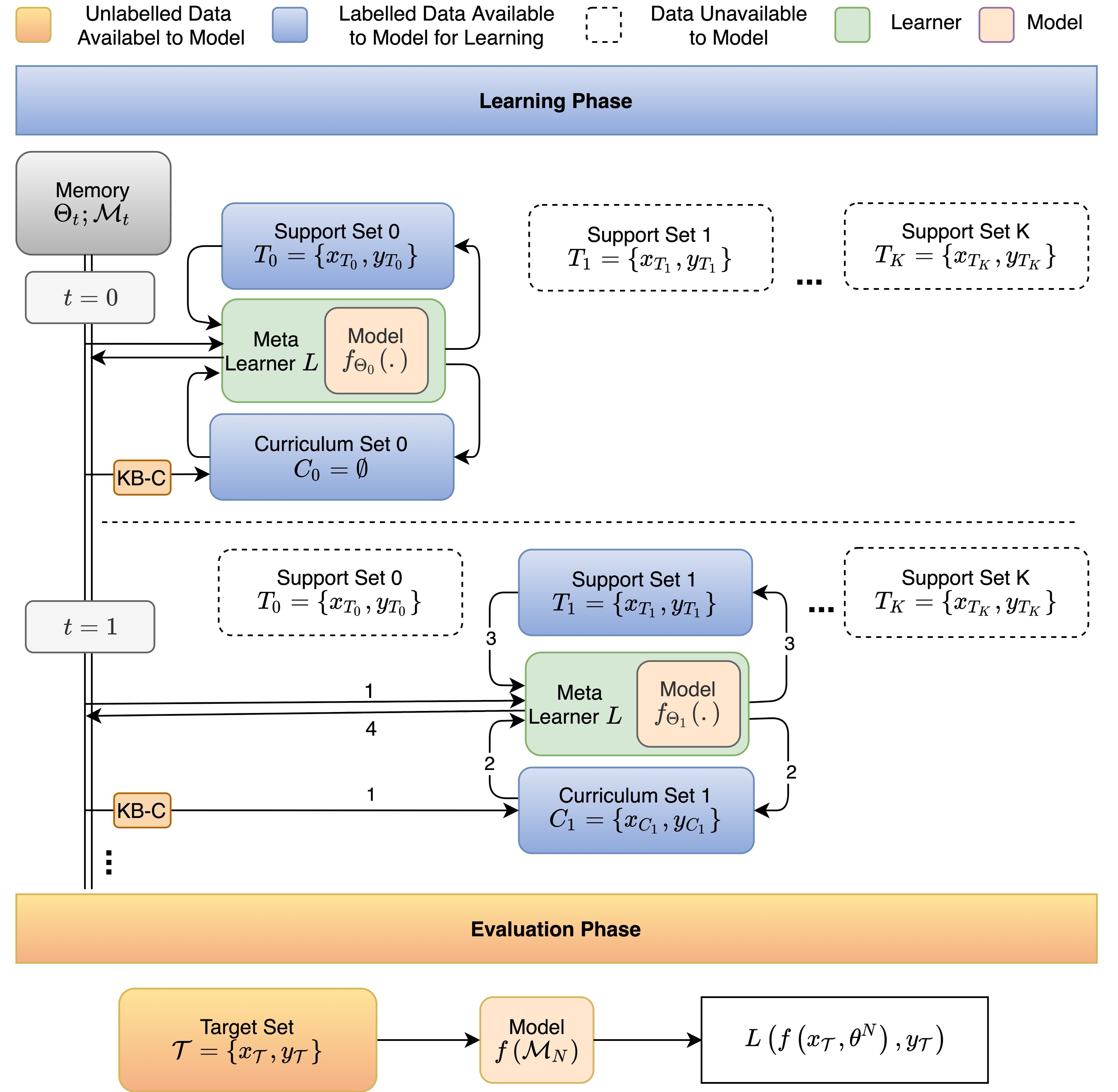}}
    \caption{The overall workflow of curriculum-meta learning in continual relation extraction. }
    \label{fig-framework}
\end{figure}

\section*{B. Configuration}

For EA-EMR, we follow the hyper-parameter settings  reported in the original paper~\cite{wang_NAACL2019}, where the training epoches is set to 3, the learning rate of the basic Vanilla model  is $lr=1e^{-3}$, the hidden size $h=200$, the batch size $bs=50$. The memory size for both EA-EMR and MLLRE are set to $ms=50\times k$, where $k$ is the number of the trained tasks. For EA-EMR, the learning rate $lr=1e^{-4}$, and $epoch=20$ for training on Continual-FewRel, and $epoch=20$ for training on Continual-SimpleQuestions. For MLLRE and CML, we follow the hyper-parameter settings of EA-EMR, and set the fixed step size $\epsilon=0.4$ and the learning rate $lr=2e^{-3}$.

For our knowledge representation model KB-C, we first utilize the TransE \cite{TransE} model to pre-train 200-dimensional entity and relation embeddings, with the learning rate $lr = 1e^{-3}$, $Margin = 1.0$, and negative sampling rate $n_r=1$. In KB-C, The dimension of the first layer is set to $d_1=50$ and second hidden layer to $d_2 = 50$, the learning rate $lr = 5e^{-4}$. 

\section*{C. Order Analysis}

In this section, we conduct a case study on Continual-Fewrel to analyze the order-sensitivity problem. For example, in Table~\ref{EMAR-roll}, $taskID = \{T_0, ..., T_9\}$ represents the fixed ID of each task, $runID=\{0, ..., 9\}$ represents the runs with loop offset from 0 to 9, $Position={P_0, ..., P_9}$ represents the position of a task in a run. $\mu$ and $\delta$ are the mean and variance of accuracy (read vertically) respectively. Different runs are distinguished by color in the table (read diagnally), and the same color represents the same round of experiment, for example, when $T_0$ is at $P_1$ and $runID=9$, we can obtain the $acc_0^1 = 71.2\%$ at the final time step, and $Acc_a = 75.8\%$, $Acc_w=55.3\%$.
\input{tab-perm}


\bibliographystyle{aaai21}
\bibliography{bibfile}

%% file: sec1-intro.tex

\section{Introduction}
Relation extraction \cite{RE_survey20} aims at extracting structured facts as triples from unstructured text. As an essential component of information extraction, relation extraction has been widely utilized in downstream applications such as knowledge base construction \cite{Wikidata_dataset} and population \cite{Yago3}. However, given the continuous and iterative nature of the update process, continual relation extraction \cite{wang_NAACL2019,MLLRE} is a more realistic and useful setting. Yet due to the limitations of storage and computational resources, it is impractical to grant the relation extractor access to all the training instances in previously seen tasks. Thus, this \emph{continual learning} formulation is in contrast to the conventional relation extraction setting where the extractor is generally trained from scratch with the full access to the training corpus. 

Catastrophic Forgetting (CF)  is a well-known problem in continual learning \cite{10.5555/3086758}. The problem is that when a neural network is utilized to learn a sequence of tasks, the learning of the later tasks may degrade the performance of the learned model for the previous tasks. 
Various recent works tackle the CF problem, including consolidation-based methods \cite{continual_cons17-2, EWC}, dynamic architecture methods \cite{continual_archi16, continual_archi16-2}, or memory-based methods \cite{continual_mem17,GEM,AGEM}. These methods have been demonstrated in simple image classiﬁcation tasks. Yet the memory-based methods have been proven to be the most promising for NLP applications. EA-EMR~\cite{wang_NAACL2019} proposes a sentence embedding alignment mechanism in memory maintenance and adopt it to continual relation extraction learning. Based on EA-EMR, MLLRE~\cite{MLLRE} introduces a meta-learning framework for fast adaptation and  EMAR~\cite{ACL20_continualRE} introduces a multi-turn co-training procedure for memory-consolidation. Most of these methods explore 
CF problem in the overall performance of task sequences, but they lack the insight analysis of the characteristics of each subtask and the corresponding model performance. 
%


Order-sensitivity (OS) is another major problem in continual learning, which is relatively under-explored \cite{10.5555/3086758,Yoon2020ScalableAO}.
It refers to the phenomenon that the performance of tasks varies based on the order of the task arrival sequence. 
This is due to not only the CF incurred by the different sequences of previous tasks but also the unidirectional knowledge transfer from the previous tasks. 
Order-sensitivity can be problematic in various aspects: (i) ethical AI considerations in continual learning, e.g.\ fairness in the medical domain \cite{Yoon2020ScalableAO}; (ii) bench-marking of continual learning algorithms as most of the existing works pick an arbitrary and random sequence of the given tasks for evaluation \cite{10.5555/3086758}; (iii) uncertainty to the quality of extracted knowledge in the realistic scenario for knowledge base population, where the model is faced with only one sequence. 

In this paper, we introduce the curriculum-meta learning (CML) method to tackle both the catastrophic forgetting and order-sensitivity problems. 
Taking a memory-based approach, CML is based on the following observations about the catastrophic forgetting and order sensitivity issues of the previous works: (i) over-fitting to the experience memory, indicating that the performance on any task will decrease as training progresses, and (ii) the interference between similar tasks, indicating that the model performs better on less intrusive tasks. 
We therefore design a mechanism which selectively reduces the replay frequency of memory to avoid over-fitting, and steer the model to learn the bias between the current task and previous most similar tasks to reduce the order-sensitivity. 
%

Our CML method contains two steps. In the first step, it samples instances from the memory based on the difficulty of the previous tasks for the current task, resulting in a curriculum for continual learning. Then, it trains the model on both the curriculum and training instances of the current task. 
We further introduce a knowledge-based method to quantify task difficulty according to the similarity of pairs of relations. Taking a relation as a function mapping to named entities in its domain and range, we define a similarity measure between two relations based on the conceptual distribution of their head and tail entities. 

Our contributions are summarized as follows: 

\begin{itemize}
\item We propose a novel curriculum-meta learning method to tackle the order-sensitivity and catastrophic forgetting problems in continual relation extraction. 
\item We introduce a new relation representation learning method via the conceptual distribution of head and tail entities of relations, which is utilized to quantify the difficulty of each relation extraction task for constructing the curriculum. 
\item We conduct comprehensive experiments to analyze the order-sensitivity and catastrophic forgetting problems in state-of-the-art methods, and empirically demonstrate that our proposed method outperforms the state-of-the-art methods on three benchmark datasets.
\end{itemize}

%% file: sec2-related.tex
\section{Related Work}

The conventional relation extraction methods could be categorized into three domains by the way data is used:  supervised methods~\cite{SRE02,SRE13,SRE14,SRE16-2,SRE16},  semi-supervised methods~\cite{semiRE06,semiRE11,semiRE19}, and distantly supervised methods~\cite{relDIS11,relDIS16}.  Most of these methods assume a predefined relation schema and thus cannon be easily generalized to new relations. 
To overcome this problem, several challenging tasks, including open relation learning and continual relation learning, have been proposed to detect and learn relations without a predefined relation schema. 

In this paper, we address the continual relation learning problem \cite{wang_NAACL2019}, a relatively new and less investigated task. Continual learning in general faces two major challenges: catastrophic forgetting and order-sensitivity. 

Catastrophic forgetting (CF) is a prominent line of research in continual learning \cite{10.5555/3086758,Thrun98}.  Methods addressing CF can be broadly divided into three categories. (i) Consolidation-based methods \cite{EWC,continual_cons17-2,continual_cons18,continual_cons18_2} consolidate model parameters important to previous tasks and reduce their learning weights. These methods employ sophisticated mechanisms to evaluate parameter importance for tasks. (ii) Dynamic architecture methods \cite{arc_ll,arc_ll2} dynamically expand model architectures to learn new tasks and effectively prevent forgetting of old tasks. Sizes of these methods grow dramatically with new tasks, making them unsuitable for NLP applications. (iii) Memory-based methods \cite{GEM,continual_mem17,continual_mem17-2,continual_mem18,AGEM} remember a few examples in old tasks and continually learn them with emerging new tasks to alleviate catastrophic forgetting. Among these methods, the memory-based methods have been proven to be the most promising for NLP applications~\cite{NLP_app20,NLP_app19}, including both relation learning~\cite{ACL20_continualRE,wang_NAACL2019}.

Order-sensitivity (OS) \cite{10.5555/3086758,Yoon2020ScalableAO} is another major problem in continual learning that is relatively under-explored. It is the phenomenon that a model's performance is sensitive to the order in which tasks arrive. In this paper, we tackle this problem by leveraging a curriculum learning method \cite{cur-l2009}. Briefly, we construct our curriculum by the similarity of tasks, thus minimizing the impact and interference of previous tasks.

%% file: sec4-framework.tex
\section{Curriculum-Meta Learning}\label{sec:method}

\subsubsection{Problem Formulation.} 
In continual relation extraction,  given a sequence of $K$ tasks $\{\mathcal{T}_1, \mathcal{T}_2, \ldots, \mathcal{T}_K\}$, each task $\mathcal{T}_k$ is a conventional supervised classification task, containing a series of examples and their corresponding labels $\{(x^{(i)}, y^{(i)})\}$, where $x^{(i)}$ is the input data, containing the natural-language context and the candidate relations, and $y^{(i)}$ is the ground-truth relation label of the context. 
The model $f_\theta(.)$ can access the training data of the current task $\mathcal{D}^{train}_k$ and is trained by optimising a loss function  $l(f_\theta(x), y)$. 
The goal of continual learning is to train the model $f_\theta(.)$ such that it continually learns new tasks while avoiding catastrophically forgetting the previously learned tasks. Due to various constraints, the learner is typically allowed to maintain and observe only a subset of the training data of the previous tasks, which is contained in a memory set $\mathcal{M}$. 

The performance of the model $f_\theta(.)$ is measured in the conventional way, by \emph{whole accuracy} $Acc_w = acc_{f, \mathcal{D}^{test}}$, on the entire test set, where $\mathcal{D}^{test}=\bigcup_{i=1}^K\mathcal{D}_i^{test}$.
Moreover, model performance at task $k$ is evaluated with \emph{average accuracy} on the test sets of all the tasks up to this task in the sequence $ Acc_a = \frac{1}{k}\sum^k_{i=1}acc_{f, i}$. Average accuracy is a better measure of the effect of catastrophic forgetting as it emphasizes on a model's performance of earlier tasks. 

\subsubsection{Framework.}
\begin{figure}[!ht]
\small
\removelatexerror
\begin{algorithm}[H]
\caption{Curriculum-Meta Learning}
\label{alg-KCAL}
\KwIn{Stream of incoming tasks $\mathcal{T}_1, \mathcal{T}_2, ..$; Classification model $f_\theta$, Step size $\epsilon$, Learning rate $\alpha$, Relations embedding $r$, Memory buffer $\mathcal{M}$, Curriculum size $k$, Curriculum instance size $n$, Knowledge-based curriculum module $g_\phi$}

Initialize $\theta$

\While {there are still tasking} {

    Retrieve current task $\mathcal{T}_t$
    
    Initialize $\theta_t \leftarrow \theta$
    
    \While{not convergence}{
    
        \If {$\mathcal{M}$ is not empty}{
        
            
            Set $\mathcal{D}_t^{train}\leftarrow\mathcal{D}_t^{train}\cup \mathcal{M}$
        }
        
        \For{each relation $R_i$ in $\mathcal{T}_t$}{
            
            Initialize $\theta_{t_i}\leftarrow\theta_t$

            Sample $D_{t,i}^{train}$ from $D_{t}^{train}$ for $R_i$
            
            \If{$\mathcal{T}_t$ is not the first task}{
        
                Sample $k$ curriculum relations from $\mathcal{M}$ with $g_\phi$.
            
                Sample $n$ instances for each curriculum relation
            
                Construct sorted mini-batches, each contains $D_{t,i}^{curri}$ of $k \times n$ instances.
                
                Evaluate $\nabla_{\theta_{t_i}}L_{R_i}(f_{\theta_{t_i}})$ using  $D_{t,i}^{curri}$
            
                Update $\theta_{t_i}' = Adam(\theta_{t_i}, L_{R_i}(f_{\theta_{t_i}}), \alpha)$
            }
            
            Evaluate $\nabla_{\theta_{t_i}'}L_{R_i}(f_{\theta_{t_i}'})$ using  $D_{t,i}^{train}$
            
            Update $\theta_{t_i}^* = Adam(\theta_{t_i}', L_{R_i}(f_{\theta_{t_i}'}), \alpha)$
            
        }
    }
    
    Update $\theta_{t+1} = \theta_t + \frac{\epsilon}{N}\sum^N_{i=1}(\theta_{t_i}^* - \theta_t)$
    
    Sample $sub\mathcal{T}_t$ from $\mathcal{T}_t$
    
    Update $\mathcal{M} \leftarrow \mathcal{M} \cup sub\mathcal{T}_t$ 
    
    Update $\theta\leftarrow\theta_t$

Fine-tune $\theta$ on $\mathcal{M}$
}
\end{algorithm}
\end{figure}
Our curriculum-meta learning (CML) framework is described in Algorithm~\ref{alg-KCAL}. CML  maintains initialization parameters $\theta_t$ and a memory set $\mathcal{M}$ that stores the prototype instances of previous tasks. 
It performs the following operations at each time step $t$ during the learning phase. (1) The meta-learner $L$ fetches the initialization parameters $\theta_t$ from the memory to initialize the model $f_{\theta_t}(.)$. (2) $L$ replays on the curriculum set $D_{t,i}^{curri}$ which is sampled and sorted by the knowledge-based curriculum module. (3) $L$ trains on the support set $D_{t,i}^{train}$ of the current task $\mathcal{T}_t$. (4) Finally, $L$ updates the learned parameters $\theta_{t+1}$ and stores a small number of prototype instances of the current task into the memory. During the evaluation phase, the trained model is given a target set with labeled unseen instances from all observed tasks (See Appendix A for the workflow of CML.)

We will introduce the framework in terms of (1) the utilization of the initialization parameters (i.e.\ meta training) and (2) the utilization of the memory set (i.e.\ the curriculum-based memory replay). 

\textbf{Meta Training.}
Meta learning, or learning to learn, aims at developing algorithms to learn the generic knowledge of how to solve tasks from a given distribution of tasks. With a given basic relation extraction model \cite{HR-LSTM-basemodel} $f_\theta(.)$ parameterized by $\theta$, we employ the gradient-based meta-learning method \cite{reptile} to learn a prior initialization $\theta_t$ at each time step $t$. During adaption to a new task, the model parameters $\theta$ are quickly updated from $\theta_t$ to the task-specific $\theta_t^*$ with a few steps of gradient descent. Formally, the meta learner $L$ updates $f_\theta(.)$ that is optimized for the following objectives:
\begin{align}
&\min_\theta\mathbb{E}_{\mathcal{T}\thicksim p(\mathcal(T)}[\mathcal{L}(\theta^*)] 
     = \min_\theta\mathbb{E}_{\mathcal{T}\thicksim p(\mathcal(T)}[\mathcal{L}(\mathcal{U}(\mathcal{D_T}\theta))]
\end{align}   

where $\mathcal{D_T}$ is the training data, $\mathcal{L}$ is the loss function for task $\mathcal{T}$, and $\mathcal{U}$ is the optimizer of $f_\theta(.)$. Then, when it converges on the current task, the model will generate the initialization parameter $\theta_{t+1}$ for the next time step $t+1$:
\begin{equation}
    \theta_{t+1} = \theta + \frac{\epsilon}{n}\sum_{t=1}^{n}(\theta_t^*-\theta_t),
\end{equation}
where the $\theta_t^*$ is the updated parameter for the current task $\mathcal{T}_k$ at time step $t$, and $n$ is the number of instances which may be processed in parallel at a time step. 

\textbf{Curriculum-based Memory Replay.}
Meta learner $L$ reviews the previous tasks in an orderly way before learning the new task. Here, we denote by $g_\phi(.)$ a function to represent the teacher which prepares the curriculum for the student network (i.e.\ the relation extractor $f_\theta(.)$) for replay. Different from  conventional experience-replay based models, the teacher function needs to master three skills:
\begin{enumerate}
    \item Assessing the difficulty of tasks. When a new task arrives, this function calculates which of all observed previous tasks interferes with the current task.
    \item Sampling instances from the memory. By sampling, we can reduce the time consumption in the replay stage and alleviate the over-fitting problems caused by the high frequency of updates on the memory.
    \item Ranking the sampled instances by a certain strategy. The teacher instructs the student model to learn the bias between the current task and observed similar tasks in the most efficient way.
\end{enumerate}

We sample the memory randomly and sort the sampled instances according to the difficulty of each previous task with respect to the current task. Based on the above requirements, we implement a knowledge-based curriculum module, which is introduced in the next section.

%% file: sec5-knowledge.tex
\section{Knowledge-based Curriculum}
Intrinsically, order-sensitivity is caused by a model's inability to guarantee optimal performance for all previous tasks. However, from the perspective of experiments, order-sensitivity is closely related to the unbalanced forgetting rate (or the unbalanced difficulty) of different tasks, where we assume that task difficulty is due to the interactions between semantically similar relations in an observed task sequence. Intuitively, if the conceptual distribution of two relations is similar, these two relations tend to be expressed in similar natural language contexts, such as the relations ``\texttt{father}'' and ``\texttt{mother}''. 

\subsubsection{Semantic Embedding-based Difficulty Function.}
To formalize this intuition, we define a difficulty estimation function based on the semantic embeddings of relations in each task. Given a set of $K$ tasks $\{\mathcal{T}_1, \mathcal{T}_2, ..., \mathcal{T}_K\}$, the difficulty of task $\mathcal{T}_i$ is defined as:
\begin{equation}
    Dl_i := \frac{1}{K-1}\sum_{\substack{j=1 \\ j \neq i}}^{K}S_i^j
    \label{eq-whole_diff}
\end{equation}

$S_j^i$ is the similarity score between tasks $\mathcal{T}_i$ and $\mathcal{T}_j$, which is defined as the average similarity among relation pairs from the two tasks:
\begin{equation}
    S_i^j := \frac{1}{M\times N}\sum_{m=1}^{M}\sum_{n=1}^{N}s_m^n,
    \label{eq-Similarity}
\end{equation}
where $M$ and $N$ are the numbers of relations in each task respectively, and  $s_m^n$ calculates the Cosine similarity between the embeddings of the two relations:
    $s_m^n := \cos(emd_m, emd_n)$. 
Using $s_m^n$, we calculate the difficulty of each relation in the memory with respect to the relations in the current task, in order to sort and sample the relations stored in memory into the final curriculum. 

\subsubsection{Relation Representation Learning.}
In order to calculate the semantic embedding of each relation, inspired by \cite{relation_sim}, we introduce a knowledge- and distribution-based representation learning method. Intuitively, the representation of a relation is learned from the types of its head and tail entities. Consider a knowledge graph  $\mathcal{G}= \{(h,r,t)\in\mathcal{E\times R\times E}\}$, where $h$ and $t$ are the head and tail entities, $r$ is the relation between them, and $\mathcal{E}$ and $\mathcal{R}$ represent the sets of entities and relations respectively. We reduce the relation representation learning task into the problem of learning the conceptual distribution of each relation which is optimized based on the following objective:
\begin{equation}
\begin{aligned}
&\min_\phi\mathcal{L}(\phi;\mathcal{G}) = \\
    &\min_\phi\sum_{(h',t';r)\in \mathcal{G}}[-\log P_\phi(h'|r)-\log P_\phi(t'|r)]
\end{aligned}
\end{equation}

where $h'$ and $t'$ are the concepts (i.e.\ the hypernyms obtained from the knowledge graph) of the head and tail entities respectively. $P_\phi(h'|r) = \exp(NN_{\phi_1}(h', r))$, and $P_\phi(t'|r) = \exp(NN_{\phi_2}(t', r))$, where $NN_\phi(a, b)=MLP_\phi(\mathbf{a})^\top \mathbf{b}$ is two-layer neural network parameterized with $\phi$. Finally, we obtain two representations $emd^h_r$ and $emd^t_r$ for each relation, which indicate the conceptual-distribution of head entities and tail entities respectively. We concatenate these two embeddings to generate the final representation of the relation $emd_r := [emd^h_r ; emd^t_r]$.

%% file: sec6-expr.tex
\section{Experiments}\label{sec:experments}
In this section, we aim to empirically address the following research questions related to our contributions:\\ \textbf{RQ1}: Why and to what extent do current memory replay-based approaches suffer from catastrophic forgetting and order-sensitivity? \\
\textbf{RQ2}: How to qualitatively and quantitatively understand task difficulty? \\
\textbf{RQ3}: Compared with the state-of-the-art methods, can our method (curriculum-meta learning with the knowledge-based curriculum)  effectively alleviate catastrophic forgetting and order-sensitivity?\\

\noindent \textbf{Datasets.}
We conduct our experiments on three datasets, including Continual-FewRel, Continual-SimpleQuestions, and Continual-TACRED, which were introduced in~\cite{ACL20_continualRE}. FewRel \cite{fewrel_dataset} is a labelled dataset which contains 80 relations and 700 instances per relation. 
SimpleQuestions is a knowledge-based question answering dataset containing single-relation questions~\cite{simplequestion-dataset}, from which a relation extraction dataset was extracted~\cite{HR-LSTM-basemodel}. The relation extraction dataset contains 1,785 relations and 72,238 training instances. TACRED~\cite{TACRED} is a well-constructed RE dataset that contains 42 relations and 21,784 examples. Considering the special relation ``n/a'' (i.e, not available) in TACRED, we follow~\cite{ACL20_continualRE} and filter out these examples with the relation ``n/a'' and use the remaining 13,012 examples for Continual-TACRED.

Following \citet{wang_NAACL2019,MLLRE}, we partition the relations of each dataset into some groups and then consider each group of relations as a distinct task $\mathcal{T}_k$. 
We form training and testing set for each task, based on the instances in the original dataset labeled by the relations in the task. 
Following the previous work, we employ two relation partitioning methods. Firstly, the unbalanced division is based on \emph{clustering}, using the averaged word embeddings \cite{Glove} of relation names with the K-means clustering algorithm~~\cite{wang_NAACL2019}. Secondly, the random partitioning into groups with a similar number of relations ~\cite{MLLRE}. For Continual-FewRel, we partition its 80 relations into 10 distinct tasks. Similarly, we partition the 1,785 relations in Continual-SimpleQuestions into 20 disjoint tasks, so as well as partition the 41 relations in Continual-TACRED into 10 tasks.\\

\noindent \textbf{Evaluation Metrics.}
We employ the following four metrics to measure model performance. Note that the last two metrics, the average forgetting rate and the error bound, are new metrics we propose in this paper. 

\begin{description}
\item[\emph{Whole Accuracy}] of the resulting model at the end of the continual learning process on the full test sets of all  tasks,
$Acc_w := acc_{f, \mathcal{D}_{test}}.$
\item[\emph{Average Accuracy}] of the resulting model trained on task $\mathcal{T}_k$ on all the test sets of all tasks seen up to stage $k$ of the continual learning process, 
$Acc_a := \frac{1}{k}\sum^k_{i=1}acc_{f, i}.$ 
Compared to $Acc_w$, $Acc_a$ highlights the catastrophic forgetting problem. However, as we will empirically show, $Acc_a$ is subject to \emph{order-sensitivity} of the tasks sequence, and thus does not accurately measure the level of forgetting on a specific task.
\item[\emph{Average Forgetting Rate}] for task $j$ after $k$ time steps, $Fr^j_{avg}$ is a new metric to evaluate task-specific model performance on order-sensitivity. 
\begin{equation}
Fr^j_{avg}:=\frac{1}{k-1}\sum^{k-1}_{i=1}\frac{\overline{acc}^j_{i+1}-\overline{acc}^j_{i}}{\overline{acc}^j_{i}},
\label{eq-fravg}
\end{equation}
where $\overline{acc}^j_{i}$ is the model's average performance on a specific task $\mathcal{T}_j$ when it appears in the $i$th position of distinct task permutations:
\begin{equation}
\overline{acc}^j_{i} := \frac{1}{(J-1)!}\sum_{\pi \in \Pi_{[1,\ldots,J]} \text{ st } \pi_i = j} acc_{i}(\pi) 
\end{equation}
where $acc_{i}(\pi)$ is the final accuracy on task $\mathcal{T}_i$ of the model trained on the permutation $\pi$, $\Pi_{[1,..,J]}$ is the set of all permutations of the tasks $\{\mathcal{T}_1,\ldots,\mathcal{T}_J\}$, and $\pi_i$ is the index of Task $\mathcal{T}_i$ of the sequence.  We note that the number of all permutations in which task $\mathcal{T}_i$ is fixed at position $j$ is $(J-1)!$. Of course we may not be able to exactly compute $\overline{acc}^j_{f, i}$ as the size of possible tasks persmutations grows exponentially. Therefore, we estimate this quantity by Monte Carlo sampling of some permutations.
\item[\emph{Error Bound}] is a new metric to evaluate the overall model performance regarding order-sensitivity.
\begin{equation}
EB := Z_{\frac{\alpha}{2}}\times\frac{\delta}{\sqrt{n}},
\label{eq-eb}
\end{equation}
where $Z_{\frac{\alpha}{2}}$ is the confidence coefficient of confidence level $\alpha$, and $\delta$ is the standard deviation of accuracy obtained from $n$ distinct task permutations. Note that a model with a lower error bound shows better robustness and less order-sensitivity for the input sequences.
\end{description}

\noindent \textbf{Baseline Models.}
We compare our proposed CML with knowledge-based curriculum-meta learning with the following baseline models, among which Vanilla is employed as the base learner for both CML and the other models (see Appendix B for hyper-parameters.):
\begin{enumerate}
    \item Vanilla \cite{HR-LSTM-basemodel}, which is the basic model for conventional supervised relation extraction not specifically designed for the continual learning setup. 
    \item EWC \cite{EWC}, which adopts elastic weight consolidation to add special $L_2$ regularization on parameter changes. Then, EWC uses Fisher information to measure the parameter importance to old tasks, and slow down the update of those parameters important to old tasks.
    \item AGEM \cite{AGEM}, which takes the gradient on sampled memorized examples from memory as the only constraint on the optimization directions of the current task.
    \item EA-EMR \cite{wang_NAACL2019}, which maintains a memory of previous tasks to alleviate the catastrophic forgetting problem.
    \item MLLRE \cite{MLLRE}, which leverages meta-learning to improve the usage efficiency of training instances. 
    \item EMAR  \cite{ACL20_continualRE}, which introduces episodic memory activation and reconsolidation to continual relation learning. 
\end{enumerate}

\subsection{Main Results}
To evaluate the overall performance of our model CML (\textbf{RQ3}), we conduct experiments on the three datasets under both task division methods: unbalanced cluster-based task division and the uniform random task division.

\begin{table*}[]
\centering
\resizebox{14cm}{!}{
\begin{tabular}{@{}cccccccc|cc|cccc@{}}
\toprule
\multicolumn{2}{c}{}                                                                    & \multicolumn{4}{c}{Continual-FewRel}                                                                                            & \multicolumn{4}{c|}{Continual-SimpQ}                                    & \multicolumn{4}{c}{Continual-TACRED}                                    \\ \cmidrule(l){3-14} 
\multicolumn{2}{c}{\multirow{-2}{*}{}}                                                  & \multicolumn{2}{c|}{$Acc_w$}                                          & \multicolumn{2}{c|}{$Acc_a$}                            & \multicolumn{2}{c|}{$Acc_w$}       & \multicolumn{2}{c|}{$Acc_a$}       & \multicolumn{2}{c|}{$Acc_w$}              & \multicolumn{2}{c}{$Acc_a$} \\ \midrule
\multicolumn{1}{c|}{Setting}                   & \multicolumn{1}{c|}{Model}             & $Acc$                       & \multicolumn{1}{c|}{$EB$}               & $Acc$         & \multicolumn{1}{c|}{$EB$}               & $Acc$         & $EB$               & $Acc$         & $EB$               & $Acc$         & \multicolumn{1}{c|}{$EB$} & $Acc$  & $EB$               \\ \midrule
\multicolumn{1}{c|}{}                          & \multicolumn{1}{c|}{Vanilla$\ddagger$} & {\color[HTML]{000000} 16.3} & \multicolumn{1}{c|}{$\pm$4.10}          & 19.7          & \multicolumn{1}{c|}{$\pm$3.90}          & 60.3          & $\pm$2.52          & 58.3          & $\pm$2.30          & 12.0          & $\pm$3.21                    & 8.7    & $\pm$2.35          \\
\multicolumn{1}{c|}{}                          & \multicolumn{1}{c|}{EWC$\dagger$}      & 27.1                        & \multicolumn{1}{c|}{$\pm$2.32}          & 30.2          & \multicolumn{1}{c|}{$\pm$2.10}          & 67.2          & $\pm$3.16          & 59.0          & $\pm$2.20          & 14.5          & $\pm$2.51                 & 14.5   & $\pm$2.90          \\
\multicolumn{1}{c|}{}                          & \multicolumn{1}{c|}{AGEM$\dagger$}     & 36.1                        & \multicolumn{1}{c|}{$\pm$2.51}          & 42.5          & \multicolumn{1}{c|}{$\pm$2.63}          & 77.6          & $\pm$2.11          & 72.2          & $\pm$2.72          & 12.5          & $\pm$2.24                 & 16.5   & $\pm$2.20          \\
\multicolumn{1}{c|}{}                          & \multicolumn{1}{c|}{EA-EMR$\ddagger$}  & 59.8                        & \multicolumn{1}{c|}{$\pm$1.50}          & 74.8          & \multicolumn{1}{c|}{$\pm$1.30}          & 82.7          & $\pm$0.48          & 86.2          & $\pm$0.33          & 17.8          & $\pm$1.01                 & 25.4   & $\pm$1.17          \\
\multicolumn{1}{c|}{}                          & \multicolumn{1}{c|}{EMAR$\dagger$}     & 53.8                        & \multicolumn{1}{c|}{$\pm$1.30}          & 68.6          & \multicolumn{1}{c|}{$\pm$0.71}          & 80.0          & $\pm$0.83          & 76.9          & $\pm$1.39          & 42.7          & $\pm$2.92                 & \textbf{52.5}   & $\pm$1.74          \\
\multicolumn{1}{c|}{}                          & \multicolumn{1}{c|}{MLLRE}             & 56.8                        & \multicolumn{1}{c|}{$\pm$1.30}          & 70.2          & \multicolumn{1}{c|}{$\pm$0.93}          & 84.5          & $\pm$0.35          & 86.7          & $\pm$0.46          & 34.4          & $\pm$\textbf{0.49}                 & 41.2   & $\pm$1.37          \\
\multicolumn{1}{c|}{\multirow{-7}{*}{Cluster}} & \multicolumn{1}{c|}{CML (ours)}          & \textbf{60.2}               & \multicolumn{1}{c|}{\textbf{$\pm$0.71}} & \textbf{76.0} & \multicolumn{1}{c|}{\textbf{$\pm$0.24}} & \textbf{85.6} & \textbf{$\pm$0.34} & \textbf{87.5} & \textbf{$\pm$0.32} & \textbf{44.4} & $\pm$1.16                 & 49.3   & \textbf{$\pm$1.01} \\ \midrule
\multicolumn{1}{c|}{}                          & \multicolumn{1}{c|}{Vanilla$\ddagger$} & 19.1                        & \multicolumn{1}{c|}{$\pm$1.20}          & 19.3          & \multicolumn{1}{c|}{$\pm$1.30}          & 55.0          & $\pm$1.30          & 55.2          & $\pm$1.30          & 10.2          & $\pm$2.02                 & 10.4   & $\pm$2.31          \\
\multicolumn{1}{c|}{}                          & \multicolumn{1}{c|}{EWC$\dagger$}      & 30.1                        & \multicolumn{1}{c|}{$\pm$1.07}          & 30.2          & \multicolumn{1}{c|}{$\pm$1.05}          & 66.4          & $\pm$0.81          & 66.7          & $\pm$0.83          & 15.3          & $\pm$1.70                 & 15.4   & $\pm$1.79          \\
\multicolumn{1}{c|}{}                          & \multicolumn{1}{c|}{AGEM$\dagger$}     & 36.9                        & \multicolumn{1}{c|}{$\pm$0.80}          & 37.0          & \multicolumn{1}{c|}{$\pm$0.83}          & 76.4          & $\pm$1.02          & 76.7          & $\pm$1.01          & 13.4          & $\pm$1.47                 & 14.3   & $\pm$1.62          \\
\multicolumn{1}{c|}{}                          & \multicolumn{1}{c|}{EA-EMR$\ddagger$}  & 61.4                        & \multicolumn{1}{c|}{$\pm$0.81}          & 61.6          & \multicolumn{1}{c|}{$\pm$0.76}          & 83.1          & $\pm$0.41          & 83.2          & $\pm$0.47          & 27.3          & $\pm$1.01                 & 30.3   & $\pm$0.70          \\
\multicolumn{1}{c|}{}                          & \multicolumn{1}{c|}{EMAR$\dagger$}     & 62.7                        & \multicolumn{1}{c|}{$\pm$0.63}          & 62.8          & \multicolumn{1}{c|}{$\pm$0.62}          & 82.4          & $\pm$0.86          & 84.0          & $\pm$0.78          & \textbf{45.1}          & $\pm$1.48                 & \textbf{46.4}   & $\pm$2.00          \\
\multicolumn{1}{c|}{}                          & \multicolumn{1}{c|}{MLLRE}             & 59.8                        & \multicolumn{1}{c|}{$\pm$0.91}          & 59.8          & \multicolumn{1}{c|}{$\pm$0.94}          & 85.2          & $\pm$0.25          & 85.5          & $\pm$0.31          & 36.4          & \textbf{$\pm$0.66}                 & 38.0   & $\pm$\textbf{0.58}          \\
\multicolumn{1}{c|}{\multirow{-7}{*}{Random}}  & \multicolumn{1}{c|}{CML (ours)}          & \textbf{62.9}               & \multicolumn{1}{c|}{\textbf{$\pm$0.62}} & \textbf{63.0} & \multicolumn{1}{c|}{\textbf{$\pm$0.59}} & \textbf{86.5} & \textbf{$\pm$0.22} & \textbf{86.9} & \textbf{$\pm$0.28} & 43.7          & $\pm$0.83                 & 45.3   & $\pm$0.72          \\ \bottomrule
\end{tabular}
}
\caption{The average accuracy $Acc_a$ and whole accuracy $Acc_w$ with error bounds by 0.95 confidence, on the test sets of observed tasks at the final time step, where $\dagger$ and $\ddagger$ indicate the result generated from the source code provided by \cite{ACL20_continualRE}\footnotemark[1] and \cite{wang_NAACL2019}\footnotemark[2] respectively.}
\label{table_main}
\end{table*}
\footnotetext[1]{https://github.com/thunlp/ContinualRE}
\footnotetext[2]{https://github.com/hongwang600/}

\begin{table}[]
\centering
\resizebox{5.8cm}{!}{
\begin{tabular}{@{}c|c|cccc@{}}
\toprule
\multicolumn{2}{c|}{Train}        & \multicolumn{2}{c|}{100}                          & \multicolumn{1}{c|}{200} & \ \ all\ \ \ \ \   \\ \midrule
\multicolumn{2}{c|}{Memory}       & \multicolumn{1}{c|}{25} & \multicolumn{1}{c|}{50} & \multicolumn{1}{c|}{50}  & 50   \\ \midrule
\multirow{2}{*}{EA-EMR} & $Acc_a$ & 70.7                    & 75.5                    & 74.8                     & 73.9 \\
                        & $Acc_w$ & 53.2                    & 57.4                    & 59.8                     & 59.6 \\ \midrule
\multirow{2}{*}{MLLRE}  & $Acc_a$ & 68.4                    & 72.1                    & 70.2                     & 51.0 \\
                        & $Acc_w$ & 51.9                    & 57.8                    & 56.8                     & 47.3 \\ \midrule
\multirow{2}{*}{EMAR}   & $Acc_a$ & 60.1                    & 66.7                    & 68.6                     & 74.1 \\
                        & $Acc_w$ & 43.7                    & 51.2                    & 53.8                     & 57.7 \\ \midrule
\multirow{2}{*}{CML}    & $Acc_a$ & \textbf{73.6}                    & \textbf{76.4}                    & \textbf{76.0}                     & 58.0 \\
                        & $Acc_w$ & \textbf{54.7}                    & \textbf{60.3}                    & \textbf{60.2}                     & 49.1 \\ \bottomrule
\end{tabular}}

\caption{Experimental results on the impact of the amount and ratio of memory on model performance over Continual-FewRel.}
\label{table-mem}
\end{table}

The following observations can be made from Table~\ref{table_main}. (i) Our model CML achieves the best $Acc_w$ and $Acc_a$ in both settings and on the three datasets in the majority of cases. (ii) Specifically, CML achieves the best $Acc_w$ and $Acc_a$ in the two larger datasets Continual-FewRel and Continual-SimpQ. (iii) CML obtains the lowest error bounds $EB$ in the majority of cases, demonstrating better stability and lower order-sensitivity. (iv) The two task division methods produce the most prominent $Arcc_a$ differences on Continual-Fewrel for CML. when the data is evenly distributed (i.e.\ Random), CML's $Acc_a$ is significantly reduced to be almost equal to $Acc_w$ (from 76.0 to 63.0). On the other two datasets, the performance difference is much less noticeable. 

Although the three metrics $Acc_a$, $Acc_w$, and $EB$ are good measures of the overall model performance, they do not provide task-specific insights, which we will further discuss in the following subsection. 

\subsection{Analysis of Unbalanced Forgetting}
We designed another experiment to better understand the reason for catastrophic forgetting and order-sensitivity  (\textbf{RQ1}).  In this experiment, each task is assigned a fixed ID. Starting with an initial ``run'' of tasks $0, 1,\ldots,9$, we test model performance on ten different runs generated by the cyclic shift of the initial run. The results of EA-EMR on Continual-Fewrel are summarised in Table~\ref{table_loop} (See Appendix C for the result for EMAR, MLLRE, and CML.)

\begin{table*}
\centering
\resizebox{12.8cm}{!}{
\begin{tabular}{@{}c|cccccccccc|cc@{}}
\toprule
$taskID$       & $\mathcal{T}_0$               & $\mathcal{T}_1$              & $\mathcal{T}_2$              & $\mathcal{T}_3$              & $\mathcal{T}_4$               & $\mathcal{T}_5$              & $\mathcal{T}_6$              & $\mathcal{T}_7$              & $\mathcal{T}_8$              & $\mathcal{T}_9$              &                              &                              \\ \cmidrule(r){1-11}
$runID$        & \cellcolor[HTML]{F2F2F2}0     & \cellcolor[HTML]{C9C9C9}1    & \cellcolor[HTML]{DBDBDB}2    & \cellcolor[HTML]{EDEDED}3    & \cellcolor[HTML]{F4B084}4     & \cellcolor[HTML]{F8CBAD}5    & \cellcolor[HTML]{FCE4D6}6    & \cellcolor[HTML]{8EA9DB}7    & \cellcolor[HTML]{B4C6E7}8    & \cellcolor[HTML]{D9E1F2}9    & \multirow{-2}{*}{$Acc_a$}    & \multirow{-2}{*}{$Acc_w$}    \\ \midrule
$\mathcal{P}_0$ & \cellcolor[HTML]{F2F2F2}88.1  & \cellcolor[HTML]{C9C9C9}39.0 & \cellcolor[HTML]{DBDBDB}49.9 & \cellcolor[HTML]{EDEDED}59.9 & \cellcolor[HTML]{F4B084}100.0 & \cellcolor[HTML]{F8CBAD}23.8 & \cellcolor[HTML]{FCE4D6}28.2 & \cellcolor[HTML]{8EA9DB}48.5 & \cellcolor[HTML]{B4C6E7}52.6 & \cellcolor[HTML]{D9E1F2}53.9 & \cellcolor[HTML]{D9E1F2}76.8 & \cellcolor[HTML]{D9E1F2}62.7 \\
$\mathcal{P}_1$ & \cellcolor[HTML]{D9E1F2}82.2  & \cellcolor[HTML]{F2F2F2}43.5 & \cellcolor[HTML]{C9C9C9}68.8 & \cellcolor[HTML]{DBDBDB}33.9 & \cellcolor[HTML]{EDEDED}100.0 & \cellcolor[HTML]{F4B084}30.7 & \cellcolor[HTML]{F8CBAD}56.7 & \cellcolor[HTML]{FCE4D6}39.6 & \cellcolor[HTML]{8EA9DB}71.1 & \cellcolor[HTML]{B4C6E7}59.0 & \cellcolor[HTML]{B4C6E7}73.4 & \cellcolor[HTML]{B4C6E7}63.9 \\
$\mathcal{P}_2$ & \cellcolor[HTML]{B4C6E7}83.0  & \cellcolor[HTML]{D9E1F2}49.9 & \cellcolor[HTML]{F2F2F2}66.7 & \cellcolor[HTML]{C9C9C9}75.9 & \cellcolor[HTML]{DBDBDB}100.0 & \cellcolor[HTML]{EDEDED}26.7 & \cellcolor[HTML]{F4B084}57.4 & \cellcolor[HTML]{F8CBAD}74.7 & \cellcolor[HTML]{FCE4D6}58.5 & \cellcolor[HTML]{8EA9DB}64.1 & \cellcolor[HTML]{8EA9DB}72.6 & \cellcolor[HTML]{8EA9DB}61.2 \\
$\mathcal{P}_3$ & \cellcolor[HTML]{8EA9DB}91.9  & \cellcolor[HTML]{B4C6E7}48.3 & \cellcolor[HTML]{D9E1F2}76.3 & \cellcolor[HTML]{F2F2F2}77.0 & \cellcolor[HTML]{C9C9C9}100.0 & \cellcolor[HTML]{DBDBDB}23.2 & \cellcolor[HTML]{EDEDED}54.3 & \cellcolor[HTML]{F4B084}77.0 & \cellcolor[HTML]{F8CBAD}83.7 & \cellcolor[HTML]{FCE4D6}50.6 & \cellcolor[HTML]{FCE4D6}62.8 & \cellcolor[HTML]{FCE4D6}59.0 \\
$\mathcal{P}_4$ & \cellcolor[HTML]{FCE4D6}90.4  & \cellcolor[HTML]{8EA9DB}44.1 & \cellcolor[HTML]{B4C6E7}73.6 & \cellcolor[HTML]{D9E1F2}79.6 & \cellcolor[HTML]{F2F2F2}100.0 & \cellcolor[HTML]{C9C9C9}43.5 & \cellcolor[HTML]{DBDBDB}47.5 & \cellcolor[HTML]{EDEDED}69.6 & \cellcolor[HTML]{F4B084}86.7 & \cellcolor[HTML]{F8CBAD}77.6 & \cellcolor[HTML]{F8CBAD}77.7 & \cellcolor[HTML]{F8CBAD}58.4 \\
$\mathcal{P}_5$ & \cellcolor[HTML]{F8CBAD}97.0  & \cellcolor[HTML]{FCE4D6}42.8 & \cellcolor[HTML]{8EA9DB}71.2 & \cellcolor[HTML]{B4C6E7}79.9 & \cellcolor[HTML]{D9E1F2}100.0 & \cellcolor[HTML]{F2F2F2}46.0 & \cellcolor[HTML]{C9C9C9}74.9 & \cellcolor[HTML]{DBDBDB}68.5 & \cellcolor[HTML]{EDEDED}80.0 & \cellcolor[HTML]{F4B084}76.8 & \cellcolor[HTML]{F4B084}79.5 & \cellcolor[HTML]{F4B084}61.6 \\
$\mathcal{P}_6$ & \cellcolor[HTML]{F4B084}98.5  & \cellcolor[HTML]{F8CBAD}79.2 & \cellcolor[HTML]{FCE4D6}69.3 & \cellcolor[HTML]{8EA9DB}75.5 & \cellcolor[HTML]{B4C6E7}100.0 & \cellcolor[HTML]{D9E1F2}52.4 & \cellcolor[HTML]{F2F2F2}82.6 & \cellcolor[HTML]{C9C9C9}88.1 & \cellcolor[HTML]{DBDBDB}88.1 & \cellcolor[HTML]{EDEDED}80.4 & \cellcolor[HTML]{EDEDED}74.6 & \cellcolor[HTML]{EDEDED}58.7 \\
$\mathcal{P}_7$ & \cellcolor[HTML]{EDEDED}97.0  & \cellcolor[HTML]{F4B084}80.2 & \cellcolor[HTML]{F8CBAD}92.1 & \cellcolor[HTML]{FCE4D6}67.9 & \cellcolor[HTML]{8EA9DB}100.0 & \cellcolor[HTML]{B4C6E7}57.1 & \cellcolor[HTML]{D9E1F2}81.9 & \cellcolor[HTML]{F2F2F2}87.6 & \cellcolor[HTML]{C9C9C9}93.3 & \cellcolor[HTML]{DBDBDB}81.4 & \cellcolor[HTML]{DBDBDB}67.4 & \cellcolor[HTML]{DBDBDB}56.3 \\
$\mathcal{P}_8$ & \cellcolor[HTML]{DBDBDB}91.9  & \cellcolor[HTML]{EDEDED}81.5 & \cellcolor[HTML]{F4B084}91.9 & \cellcolor[HTML]{F8CBAD}92.7 & \cellcolor[HTML]{FCE4D6}99.3  & \cellcolor[HTML]{8EA9DB}68.0 & \cellcolor[HTML]{B4C6E7}86.1 & \cellcolor[HTML]{D9E1F2}93.4 & \cellcolor[HTML]{F2F2F2}92.6 & \cellcolor[HTML]{C9C9C9}90.8 & \cellcolor[HTML]{C9C9C9}77.4 & \cellcolor[HTML]{C9C9C9}56.1 \\
$\mathcal{P}_9$ & \cellcolor[HTML]{C9C9C9}100.0 & \cellcolor[HTML]{DBDBDB}89.7 & \cellcolor[HTML]{EDEDED}97.4 & \cellcolor[HTML]{F4B084}96.0 & \cellcolor[HTML]{F8CBAD}100.0 & \cellcolor[HTML]{FCE4D6}82.1 & \cellcolor[HTML]{8EA9DB}91.9 & \cellcolor[HTML]{B4C6E7}94.7 & \cellcolor[HTML]{D9E1F2}99.3 & \cellcolor[HTML]{F2F2F2}93.1 & \cellcolor[HTML]{F2F2F2}77.7 & \cellcolor[HTML]{F2F2F2}58.7 \\ \midrule
$\mu$          & 92.0                          & 59.8                         & 75.7                         & 73.8                         & 99.9                          & 45.4                         & 66.2                         & 74.2                         & 80.6                         & 72.8                         & 74.0                         & 59.7                         \\
$\sigma$       & 6.25                          & 20.06                        & 14.39                        & 17.51                        & 0.22                          & 19.92                        & 20.41                        & 18.50                        & 15.35                        & 14.99                        & 5.26                         & 2.61                         \\ \bottomrule
\end{tabular}
}
\caption{A case study of EA-EMR on the FewRel dataset, where $taskID = \{\mathcal{T}_0, ..., \mathcal{T}_9\}$ represents the fixed ID of each task, $runID=\{0, ..., 9\}$ represents the runs with loop offset from 0 to 9, $Position={\mathcal{P}_0, ..., \mathcal{P}_9}$ represents the position of a task in a run. $\mu$ and $\delta$ are the mean and variance of accuracy (read vertically) respectively. Different runs are distinguished by color in the table (read diagnally), and the same color represents the same round of experiment, for example, when $\mathcal{T}_0$ is at $\mathcal{P}_1$ and $runID=9$, we can obtain the $acc_0^1 = 82.2\%$ at the final time step, and $Acc_a = 76.8\%$, $Acc_w=62.7\%$. See Appendix C for such result of EMAR, MLLRE and the proposed CML.
}
\label{table_loop}
\end{table*}

As shown in Table~\ref{table_loop}, comparing the results in the rows of $\mathcal{P}_0$ and $\mathcal{P}_9$ of each task, we can observe that most tasks see a significant drop in accuracy from $\mathcal{P}_9$ (when the task is the last one seen by the model) to $\mathcal{P}_0$ (when the task is the first one seen by the model), indicating that they suffer from catastrophic forgetting. 

Intuitively, forgetting on a task reflects an increase in empirical error. 
We hypothesized that this is most likely due to frequent replay of limited task-related memory, in other words, over-fitting on memory. We tested this hypothesis by adjusting the ratio of training data to memory. 
Table~\ref{table-mem} shows that when memory size is fixed (e.g.\ 50), more training data (i.e.\ all vs 200 vs 100) results in poorer performance for all models except EMAR, indicating the model is over-fitting on memory. This is due to a higher replay frequency of the memory. When we fix the ratio of the training data to memory (e.g.\ 100:25 and 200:50), model performance conforms to the general rule of better performance with more data. Thus, the results shown in Table~\ref{table-mem} support our hypothesis. 

When reading each column in Table~\ref{table_loop} separately, we find that the forgetting rate of each task is different, which cannot be solely explained by the issue of over-fitting on memory. For example, the model performance on $\mathcal{T}_4$ and $\mathcal{T}_0$ is exactly the same (100\%) when they appear at the last position $\mathcal{P}_9$. However, performance on $\mathcal{T}_4$ does not degrade with the advance of its position. On the other hand, we can observe a decreasing trend of performance on $\mathcal{T}_0$ as its position moves back. 

Moreover, we observe that order-sensitivity may be related to the difficulty of the earlier task in a run, where difficulty refers to a task's tendency of being more easily forgotten by the model. For instance, we can observe that the final $Acc_a$ of the $runID=4$ experiment (in dark-orange) is the highest, whereas the final $Acc_a$ of the $runID=6$ experiment (in light orange) is the lowest. 
Also, among all tasks at position $\mathcal{P}_0$, task $\mathcal{T}_4$ has the highest accuracy while task $\mathcal{T}_6$ has the lowest. 

Both of the above observations may be explained by task difficulty, which we further study in the next subsection. 

\begin{table*}[t]
\centering
\resizebox{13cm}{!}{
\begin{tabular}{@{}c|c|cccccccccc|c@{}}
\toprule
\multicolumn{2}{c|}{}                & $\mathcal{T}_0$  & $\mathcal{T}_1$ & $\mathcal{T}_2$ & $\mathcal{T}_3$  & $\mathcal{T}_4$  & $\mathcal{T}_5$ & $\mathcal{T}_6$ & $\mathcal{T}_7$ & $\mathcal{T}_8$ & $\mathcal{T}_9$ & $PCC_s$ \\ \midrule
\multicolumn{2}{c|}{$D_{prior}$}     & 0.121  & 0.141 & 0.168 & 0.054  & 0.035  & 0.186 & 0.112 & 0.152 & 0.146 & 0.137 & -       \\ \midrule
\multirow{4}{*}{$D_{post}$} & EA-EMR & 0.060   & 0.098 & 0.061 & 0.028  & 0.001  & 0.137 & 0.139 & 0.060 & 0.044 & 0.051 & 0.559   \\
                            & MLLRE  & 0.022  & 0.078 & 0.091 & 0.085  & -0.004 & 0.147 & 0.060 & 0.064 & 0.069 & 0.075 & 0.667   \\
                            & EMAR   & 0.036 & 0.016 & 0.027 & 0.007 & 0.006 & 0.016 & 0.008 & 0.026 & 0.020 & 0.005 & 0.499  \\
                            & CML    & -0.002 & 0.108 & 0.051 & 0.065  & -0.002 & 0.113 & 0.108 & 0.046 & 0.070 & 0.068 & \textbf{0.470}   \\ \bottomrule
\end{tabular}}
\caption{Prior difficulty and posterior difficulty of each task, where $PCCs$ indicates the Pearson correlation coefficient between the estimated prior difficulty by Eq.(\ref{eq-whole_diff}) (i.e., $D_{prior}$) and the average forgetting rate of each model (i.e., $D_{post}$).}
\label{table-difficulty}
\end{table*}

\subsection{Analysis of Task Difficulty}
In this section, we present the qualitative and quantitative analyses in order to better understand the difficulty of tasks (\textbf{RQ2}). 
We choose EA-EMR as the case study on the Continual-FewRel dataset with the tasks constructed through clustering. 

\subsubsection{Qualitative Analysis.}

\begin{figure}[htbp] 
\centering
\includegraphics[width=.45\textwidth]{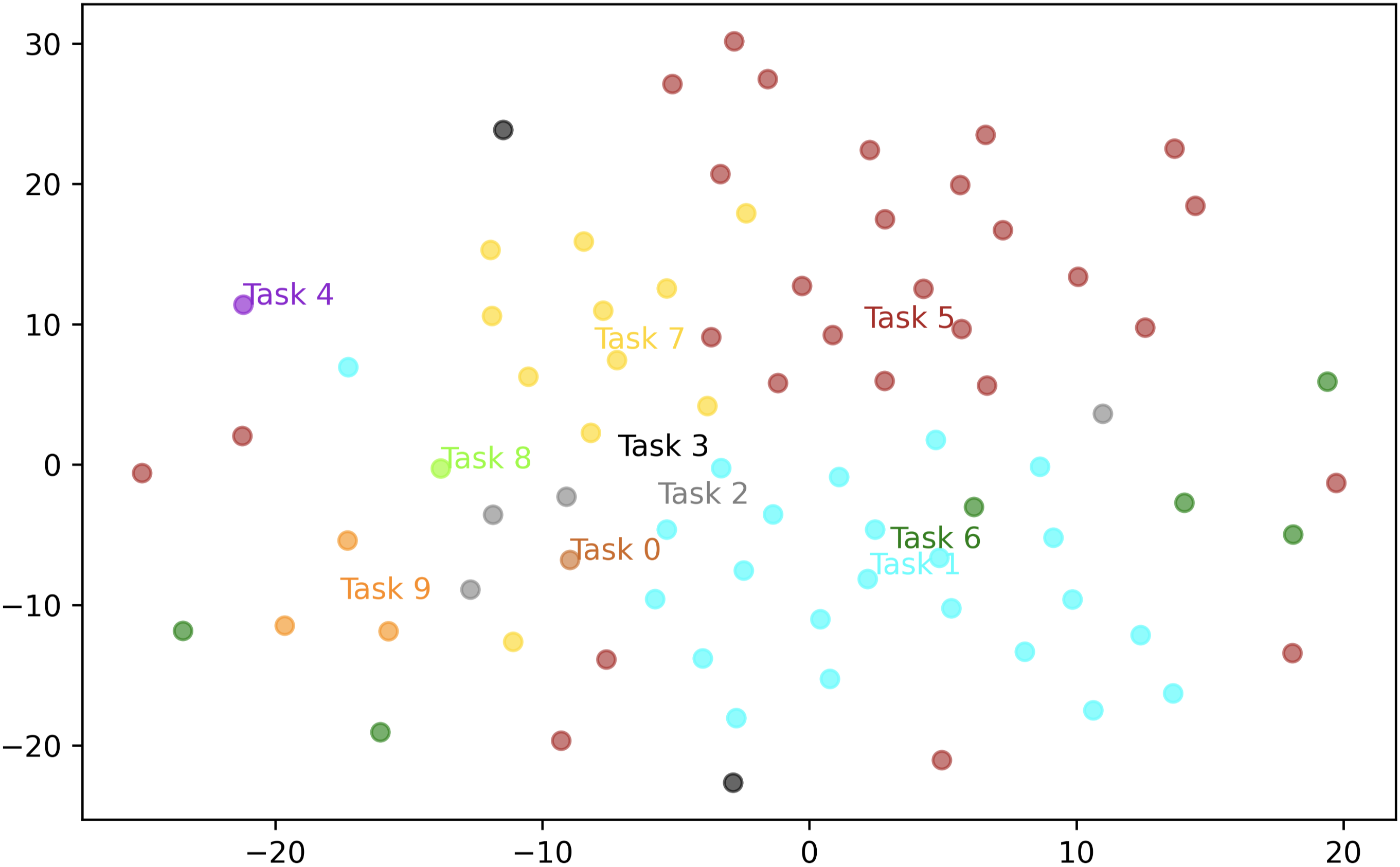}
\caption{The t-SNE \cite{tSNE} visualization of  the encoding of relations generated by the hidden layer in EA-EMR over Continual-FewRel, where each point represents a relationship, each color represents a task, and the position anchored by the label is the center of the task.}
\label{fig_plot} 
\end{figure} 


Figure~\ref{fig_plot} shows the t-SNE visualization of the relations, where nodes represent relations, colors represent the tasks, and the distance is calculated from the hidden layer in EA-EMR. 

As can be seen from the figure, tasks $\mathcal{T}_4$ and $\mathcal{T}_8$ have only one relation each, and that the relation in $\mathcal{T}_4$ is far away from the others but the relation $\mathcal{T}_8$ is much closer. The difference in their distances from the other relations may explain the difference in their task-specific performance in Table~\ref{table_loop}, where $\mathcal{T}_4$ does not suffer from catastrophic forgetting but $\mathcal{T}_8$ does. In other words, $\mathcal{T}_4$ is easy whereas $\mathcal{T}_8$ is more difficult.

Similarly, we can observe that $\mathcal{T}_3$ (colored in black) contains only two relations, and it overlaps significantly with the other tasks. Therefore, catastrophic forgetting is more serious on $\mathcal{T}_3$, i.e., $\mathcal{T}_3$ is difficult.
Finally, we can observe that tasks $\mathcal{T}_1$ and $\mathcal{T}_6$ are difficult tasks, as their centroids are very close. They both suffer from serious catastrophic forgetting as can be seen from Table~\ref{table_loop}. 

Therefore, in continual learning, the difficulty of a task $\mathcal{T}$ may be characterized by the correlation between $\mathcal{T}$ and the other observed tasks. In the relation extraction scenario, we define this correlation as the semantic similarity between the relations. 

\subsubsection{Quantitative Analysis.}
Based on the above analysis, we hypothesize that a model could achieve better performance if we can measure the similarity of the current task and previous tasks and guide the model to distinguish similar tasks. 

For ease of expression, we denote the measure difficulty of each task $\mathcal{T}_i$ as the \emph{prior difficulty} $D_{prior}$, and denote the average forgetting rate $Fr^i_{avg}$ as the \emph{posterior difficulty} $D_{post}$. The prior difficulty $D_{prior}$ is the estimated value of knowledge graph embedding, and the posterior difficulty is related to the performance of each model.

Specifically, we use our embedding-based task difficulty function defined in Eq. (\ref{eq-whole_diff}) as the prior difficulty $D_{prior}$, and average forgetting rate defined in Eq.~(\ref{eq-fravg}) as the posterior difficulty.  Table~\ref{table-difficulty} shows the correlation between the prior difficulty and the posterior difficulty, and thus offers evidence of the effectiveness of our representation learning method.

We can learn from Table~\ref{table-difficulty} four main conclusions: (i) The $PCCs$ of the three models demonstrate that the semantic embedding-based prior difficulty does positive to the forgetting rate indicating which tasks are difficult for a model. (ii) Comparing the $PCCs$ values of CML with the other three methods, our proposed CML with the knowledge-based curriculum does alleviate the interference between similar tasks as it achieves the lowest forgetting rate. (iii) For tasks $\mathcal{T}_0$ and $\mathcal{T}_4$ in CML, the average forgetting rate is negative, which means that the accuracy of these tasks decreases as they move towards the end. (iv) Based on the analysis of Table~\ref{table_main}, Table~\ref{table_loop} (See Appendix C for the other three relevant tables), it is evident that CML improves the overall accuracy and effectively alleviates the order-sensitivity problem. We note that the improvements do come at a cost of moderately decreased accuracy on simple tasks such as $\mathcal{T}_0$ and $\mathcal{T}_4$. 

%% file: sec7-conc.tex
\section{Conclusion}

In this paper, we proposed a novel curriculum-meta learning method to tackle the catastrophic forgetting and order-sensitivity issues in continual relation learning. 
The construction of the curriculum is based on the notion of task difficulty, which is defined through a novel relation representation learning method that learns from the distribution of domain and range types of relations. 
Our comprehensive experiments on the three benchmark datasets Continual-FewRel, Continual-SimpleQuestions and Continual-Tacred show that our proposed method outperforms the state-of-the-art models, and is less prone to catastrophic forgetting and less order-sensitive. In future, we will investigate an end-to-end curriculum model and a new dynamic difficulty measurement based on the framework presented in this paper.


%% file: tab-perm.tex
\begin{table*}[]
\centering
\resizebox{13cm}{!}{
\begin{tabular}{@{}c|cccccccccccc@{}}
\toprule
$taskID$       & $\mathcal{T}_0$              & $\mathcal{T}_1$              & $\mathcal{T}_2$              & $\mathcal{T}_3$              & $\mathcal{T}_4$              & $\mathcal{T}_5$              & $\mathcal{T}_6$              & $\mathcal{T}_7$              & $\mathcal{T}_8$              & \multicolumn{1}{c|}{$\mathcal{T}_9$}              &                              &                              \\ \cmidrule(r){1-11}
$runID$        & \cellcolor[HTML]{F2F2F2}0    & \cellcolor[HTML]{C9C9C9}1    & \cellcolor[HTML]{DBDBDB}2    & \cellcolor[HTML]{EDEDED}3    & \cellcolor[HTML]{F4B084}4    & \cellcolor[HTML]{F8CBAD}5    & \cellcolor[HTML]{FCE4D6}6    & \cellcolor[HTML]{8EA9DB}7    & \cellcolor[HTML]{B4C6E7}8    & \multicolumn{1}{c|}{\cellcolor[HTML]{D9E1F2}9}    & \multirow{-2}{*}{$Acc_a$}    & \multirow{-2}{*}{$Acc_w$}    \\ \midrule
$\mathcal{P}_0$ & \cellcolor[HTML]{F2F2F2}70.5 & \cellcolor[HTML]{C9C9C9}39.7 & \cellcolor[HTML]{DBDBDB}69.5 & \cellcolor[HTML]{EDEDED}90.9 & \cellcolor[HTML]{F4B084}90.5 & \cellcolor[HTML]{F8CBAD}47.5 & \cellcolor[HTML]{FCE4D6}82.3 & \cellcolor[HTML]{8EA9DB}60.1 & \cellcolor[HTML]{B4C6E7}85.2 & \multicolumn{1}{c|}{\cellcolor[HTML]{D9E1F2}93.5} & \cellcolor[HTML]{D9E1F2}75.8 & \cellcolor[HTML]{D9E1F2}55.3 \\
$\mathcal{P}_1$ & \cellcolor[HTML]{D9E1F2}71.2 & \cellcolor[HTML]{F2F2F2}46.9 & \cellcolor[HTML]{C9C9C9}79.3 & \cellcolor[HTML]{DBDBDB}88.7 & \cellcolor[HTML]{EDEDED}92.1 & \cellcolor[HTML]{F4B084}46.0 & \cellcolor[HTML]{F8CBAD}75.2 & \cellcolor[HTML]{FCE4D6}65.3 & \cellcolor[HTML]{8EA9DB}95.9 & \multicolumn{1}{c|}{\cellcolor[HTML]{B4C6E7}75.1} & \cellcolor[HTML]{B4C6E7}73.4 & \cellcolor[HTML]{B4C6E7}57.8 \\
$\mathcal{P}_2$ & \cellcolor[HTML]{B4C6E7}76.7 & \cellcolor[HTML]{D9E1F2}50.3 & \cellcolor[HTML]{F2F2F2}80.4 & \cellcolor[HTML]{C9C9C9}91.3 & \cellcolor[HTML]{DBDBDB}93.7 & \cellcolor[HTML]{EDEDED}50.6 & \cellcolor[HTML]{F4B084}76.3 & \cellcolor[HTML]{F8CBAD}65.9 & \cellcolor[HTML]{FCE4D6}79.5 & \multicolumn{1}{c|}{\cellcolor[HTML]{8EA9DB}72.9} & \cellcolor[HTML]{8EA9DB}75.7 & \cellcolor[HTML]{8EA9DB}55.6 \\
$\mathcal{P}_3$ & \cellcolor[HTML]{8EA9DB}71.9 & \cellcolor[HTML]{B4C6E7}49.0 & \cellcolor[HTML]{D9E1F2}78.0 & \cellcolor[HTML]{F2F2F2}90.2 & \cellcolor[HTML]{C9C9C9}90.5 & \cellcolor[HTML]{DBDBDB}54.7 & \cellcolor[HTML]{EDEDED}84.6 & \cellcolor[HTML]{F4B084}53.5 & \cellcolor[HTML]{F8CBAD}70.5 & \multicolumn{1}{c|}{\cellcolor[HTML]{FCE4D6}72.4} & \cellcolor[HTML]{FCE4D6}73.4 & \cellcolor[HTML]{FCE4D6}58.0 \\
$\mathcal{P}_4$ & \cellcolor[HTML]{FCE4D6}84.2 & \cellcolor[HTML]{8EA9DB}45.7 & \cellcolor[HTML]{B4C6E7}76.4 & \cellcolor[HTML]{D9E1F2}89.8 & \cellcolor[HTML]{F2F2F2}95.2 & \cellcolor[HTML]{C9C9C9}57.4 & \cellcolor[HTML]{DBDBDB}67.4 & \cellcolor[HTML]{EDEDED}58.9 & \cellcolor[HTML]{F4B084}84.4 & \multicolumn{1}{c|}{\cellcolor[HTML]{F8CBAD}76.6} & \cellcolor[HTML]{F8CBAD}74.0 & \cellcolor[HTML]{F8CBAD}57.8 \\
$\mathcal{P}_5$ & \cellcolor[HTML]{F8CBAD}82.9 & \cellcolor[HTML]{FCE4D6}56.7 & \cellcolor[HTML]{8EA9DB}76.9 & \cellcolor[HTML]{B4C6E7}95.3 & \cellcolor[HTML]{D9E1F2}96.0 & \cellcolor[HTML]{F2F2F2}47.8 & \cellcolor[HTML]{C9C9C9}72.6 & \cellcolor[HTML]{DBDBDB}51.4 & \cellcolor[HTML]{EDEDED}82.0 & \multicolumn{1}{c|}{\cellcolor[HTML]{F4B084}78.1} & \cellcolor[HTML]{F4B084}73.6 & \cellcolor[HTML]{F4B084}58.5 \\
$\mathcal{P}_6$ & \cellcolor[HTML]{F4B084}71.2 & \cellcolor[HTML]{F8CBAD}51.5 & \cellcolor[HTML]{FCE4D6}78.9 & \cellcolor[HTML]{8EA9DB}92.0 & \cellcolor[HTML]{B4C6E7}88.1 & \cellcolor[HTML]{D9E1F2}50.5 & \cellcolor[HTML]{F2F2F2}76.3 & \cellcolor[HTML]{C9C9C9}55.0 & \cellcolor[HTML]{DBDBDB}91.0 & \multicolumn{1}{c|}{\cellcolor[HTML]{EDEDED}79.9} & \cellcolor[HTML]{EDEDED}71.5 & \cellcolor[HTML]{EDEDED}58.6 \\
$\mathcal{P}_7$ & \cellcolor[HTML]{EDEDED}89.0 & \cellcolor[HTML]{F4B084}51.9 & \cellcolor[HTML]{F8CBAD}90.9 & \cellcolor[HTML]{FCE4D6}89.1 & \cellcolor[HTML]{8EA9DB}96.0 & \cellcolor[HTML]{B4C6E7}45.1 & \cellcolor[HTML]{D9E1F2}69.8 & \cellcolor[HTML]{F2F2F2}55.1 & \cellcolor[HTML]{C9C9C9}89.3 & \multicolumn{1}{c|}{\cellcolor[HTML]{DBDBDB}80.4} & \cellcolor[HTML]{DBDBDB}73.7 & \cellcolor[HTML]{DBDBDB}59.0 \\
$\mathcal{P}_8$ & \cellcolor[HTML]{DBDBDB}83.6 & \cellcolor[HTML]{EDEDED}57.2 & \cellcolor[HTML]{F4B084}76.5 & \cellcolor[HTML]{F8CBAD}92.4 & \cellcolor[HTML]{FCE4D6}88.1 & \cellcolor[HTML]{8EA9DB}47.1 & \cellcolor[HTML]{B4C6E7}77.8 & \cellcolor[HTML]{D9E1F2}48.1 & \cellcolor[HTML]{F2F2F2}75.4 & \multicolumn{1}{c|}{\cellcolor[HTML]{C9C9C9}87.7} & \cellcolor[HTML]{C9C9C9}73.6 & \cellcolor[HTML]{C9C9C9}56.3 \\
$\mathcal{P}_9$ & \cellcolor[HTML]{C9C9C9}87.0 & \cellcolor[HTML]{DBDBDB}37.6 & \cellcolor[HTML]{EDEDED}78.7 & \cellcolor[HTML]{F4B084}93.8 & \cellcolor[HTML]{F8CBAD}92.9 & \cellcolor[HTML]{FCE4D6}48.2 & \cellcolor[HTML]{8EA9DB}80.1 & \cellcolor[HTML]{B4C6E7}63.9 & \cellcolor[HTML]{D9E1F2}87.7 & \multicolumn{1}{c|}{\cellcolor[HTML]{F2F2F2}88.4} & \cellcolor[HTML]{F2F2F2}73.0 & \cellcolor[HTML]{F2F2F2}54.6 \\ \midrule
$\mu$          & 78.8                         & 48.7                         & 78.6                         & 91.4                         & 92.3                         & 49.5                         & 76.2                         & 57.7                         & 84.1                         & \multicolumn{1}{c|}{80.5}                         & 73.8                         & 57.2                         \\
$\sigma$       & 7.28                         & 6.44                         & 5.27                         & 2.09                         & 2.98                         & 3.91                         & 5.35                         & 6.09                         & 7.56                         & 7.13                                              & 1.25                         & 1.56                         \\ \bottomrule
\end{tabular}
}
\caption{A case study of EMAR~\cite{ACL20_continualRE} on the FewRel dataset.}
\label{EMAR-roll}
\end{table*}

\begin{table*}
\centering
\resizebox{13cm}{!}{
\begin{tabular}{@{}c|cccccccccc|cc@{}}
\toprule
$taskID$       & $\mathcal{T}_0$              & $\mathcal{T}_1$              & $\mathcal{T}_2$              & $\mathcal{T}_3$              & $\mathcal{T}_4$               & $\mathcal{T}_5$              & $\mathcal{T}_6$              & $\mathcal{T}_7$              & $\mathcal{T}_8$              & $\mathcal{T}_9$              &                              &                              \\ \cmidrule(r){1-11}
$runID$        & \cellcolor[HTML]{F2F2F2}0    & \cellcolor[HTML]{C9C9C9}1    & \cellcolor[HTML]{DBDBDB}2    & \cellcolor[HTML]{EDEDED}3    & \cellcolor[HTML]{F4B084}4     & \cellcolor[HTML]{F8CBAD}5    & \cellcolor[HTML]{FCE4D6}6    & \cellcolor[HTML]{8EA9DB}7    & \cellcolor[HTML]{B4C6E7}8    & \cellcolor[HTML]{D9E1F2}9    & \multirow{-2}{*}{$Acc_a$}    & \multirow{-2}{*}{$Acc_w$}    \\ \midrule
$\mathcal{P}_0$ & \cellcolor[HTML]{F2F2F2}83.0 & \cellcolor[HTML]{C9C9C9}44.1 & \cellcolor[HTML]{DBDBDB}48.8 & \cellcolor[HTML]{EDEDED}53.6 & \cellcolor[HTML]{F4B084}100.0 & \cellcolor[HTML]{F8CBAD}25.1 & \cellcolor[HTML]{FCE4D6}49.8 & \cellcolor[HTML]{8EA9DB}52.7 & \cellcolor[HTML]{B4C6E7}63.0 & \cellcolor[HTML]{D9E1F2}54.7 & \cellcolor[HTML]{D9E1F2}74.8 & \cellcolor[HTML]{D9E1F2}59.7 \\
$\mathcal{P}_1$ & \cellcolor[HTML]{D9E1F2}91.9 & \cellcolor[HTML]{F2F2F2}39.0 & \cellcolor[HTML]{C9C9C9}51.0 & \cellcolor[HTML]{DBDBDB}62.4 & \cellcolor[HTML]{EDEDED}100.0 & \cellcolor[HTML]{F4B084}25.6 & \cellcolor[HTML]{F8CBAD}53.1 & \cellcolor[HTML]{FCE4D6}60.7 & \cellcolor[HTML]{8EA9DB}48.1 & \cellcolor[HTML]{B4C6E7}53.9 & \cellcolor[HTML]{B4C6E7}75.3 & \cellcolor[HTML]{B4C6E7}60.3 \\
$\mathcal{P}_2$ & \cellcolor[HTML]{B4C6E7}94.1 & \cellcolor[HTML]{D9E1F2}43.5 & \cellcolor[HTML]{F2F2F2}62.3 & \cellcolor[HTML]{C9C9C9}70.4 & \cellcolor[HTML]{DBDBDB}100.0 & \cellcolor[HTML]{EDEDED}23.8 & \cellcolor[HTML]{F4B084}56.2 & \cellcolor[HTML]{F8CBAD}75.3 & \cellcolor[HTML]{FCE4D6}72.6 & \cellcolor[HTML]{8EA9DB}44.0 & \cellcolor[HTML]{8EA9DB}71.0 & \cellcolor[HTML]{8EA9DB}60.9 \\
$\mathcal{P}_3$ & \cellcolor[HTML]{8EA9DB}84.4 & \cellcolor[HTML]{B4C6E7}49.7 & \cellcolor[HTML]{D9E1F2}72.8 & \cellcolor[HTML]{F2F2F2}62.4 & \cellcolor[HTML]{C9C9C9}100.0 & \cellcolor[HTML]{DBDBDB}31.1 & \cellcolor[HTML]{EDEDED}58.9 & \cellcolor[HTML]{F4B084}71.5 & \cellcolor[HTML]{F8CBAD}74.2 & \cellcolor[HTML]{FCE4D6}62.6 & \cellcolor[HTML]{FCE4D6}72.4 & \cellcolor[HTML]{FCE4D6}62.6 \\
$\mathcal{P}_4$ & \cellcolor[HTML]{FCE4D6}97.0 & \cellcolor[HTML]{8EA9DB}46.1 & \cellcolor[HTML]{B4C6E7}62.3 & \cellcolor[HTML]{D9E1F2}80.7 & \cellcolor[HTML]{F2F2F2}100.0 & \cellcolor[HTML]{C9C9C9}31.7 & \cellcolor[HTML]{DBDBDB}63.2 & \cellcolor[HTML]{EDEDED}68.0 & \cellcolor[HTML]{F4B084}73.4 & \cellcolor[HTML]{F8CBAD}73.0 & \cellcolor[HTML]{F8CBAD}71.6 & \cellcolor[HTML]{F8CBAD}59.0 \\
$\mathcal{P}_5$ & \cellcolor[HTML]{F8CBAD}96.6 & \cellcolor[HTML]{FCE4D6}63.3 & \cellcolor[HTML]{8EA9DB}59.5 & \cellcolor[HTML]{B4C6E7}75.5 & \cellcolor[HTML]{D9E1F2}100.0 & \cellcolor[HTML]{F2F2F2}45.2 & \cellcolor[HTML]{C9C9C9}68.0 & \cellcolor[HTML]{DBDBDB}80.4 & \cellcolor[HTML]{EDEDED}70.4 & \cellcolor[HTML]{F4B084}72.5 & \cellcolor[HTML]{F4B084}76.1 & \cellcolor[HTML]{F4B084}57.7 \\
$\mathcal{P}_6$ & \cellcolor[HTML]{F4B084}95.6 & \cellcolor[HTML]{F8CBAD}64.4 & \cellcolor[HTML]{FCE4D6}71.2 & \cellcolor[HTML]{8EA9DB}59.9 & \cellcolor[HTML]{B4C6E7}99.3  & \cellcolor[HTML]{D9E1F2}45.3 & \cellcolor[HTML]{F2F2F2}77.7 & \cellcolor[HTML]{C9C9C9}77.2 & \cellcolor[HTML]{DBDBDB}84.4 & \cellcolor[HTML]{EDEDED}70.7 & \cellcolor[HTML]{EDEDED}74.7 & \cellcolor[HTML]{EDEDED}57.6 \\
$\mathcal{P}_7$ & \cellcolor[HTML]{EDEDED}95.6 & \cellcolor[HTML]{F4B084}76.4 & \cellcolor[HTML]{F8CBAD}80.5 & \cellcolor[HTML]{FCE4D6}59.1 & \cellcolor[HTML]{8EA9DB}98.6  & \cellcolor[HTML]{B4C6E7}52.4 & \cellcolor[HTML]{D9E1F2}81.1 & \cellcolor[HTML]{F2F2F2}80.7 & \cellcolor[HTML]{C9C9C9}92.6 & \cellcolor[HTML]{DBDBDB}70.5 & \cellcolor[HTML]{DBDBDB}73.6 & \cellcolor[HTML]{DBDBDB}59.7 \\
$\mathcal{P}_8$ & \cellcolor[HTML]{DBDBDB}91.9 & \cellcolor[HTML]{EDEDED}77.6 & \cellcolor[HTML]{F4B084}85.6 & \cellcolor[HTML]{F8CBAD}75.3 & \cellcolor[HTML]{FCE4D6}97.2  & \cellcolor[HTML]{8EA9DB}68.5 & \cellcolor[HTML]{B4C6E7}77.1 & \cellcolor[HTML]{D9E1F2}81.2 & \cellcolor[HTML]{F2F2F2}85.9 & \cellcolor[HTML]{C9C9C9}86.8 & \cellcolor[HTML]{C9C9C9}74.9 & \cellcolor[HTML]{C9C9C9}56.2 \\
$\mathcal{P}_9$ & \cellcolor[HTML]{C9C9C9}96.3 & \cellcolor[HTML]{DBDBDB}73.9 & \cellcolor[HTML]{EDEDED}92.6 & \cellcolor[HTML]{F4B084}88.3 & \cellcolor[HTML]{F8CBAD}98.2  & \cellcolor[HTML]{FCE4D6}70.6 & \cellcolor[HTML]{8EA9DB}78.7 & \cellcolor[HTML]{B4C6E7}84.1 & \cellcolor[HTML]{D9E1F2}93.3 & \cellcolor[HTML]{F2F2F2}88.0 & \cellcolor[HTML]{F2F2F2}73.2 & \cellcolor[HTML]{F2F2F2}55.6 \\ \midrule
$\mu$          & 92.6                         & 57.8                         & 68.7                         & 68.8                         & 99.3                          & 41.9                         & 66.4                         & 73.2                         & 75.8                         & 67.7                         & 73.8                         & 58.9                         \\
$\sigma$       & 5.05                         & 14.98                        & 14.51                        & 11.04                        & 1.00                          & 17.51                        & 11.71                        & 10.10                        & 13.92                        & 14.11                        & 1.69                         & 2.17                         \\ \bottomrule
\end{tabular}
}
\caption{A case study of MLLRE~\cite{MLLRE} on the FewRel dataset.}
\label{MLLRE-roll}
\end{table*}

\begin{table*}
\centering
\resizebox{13cm}{!}{
\begin{tabular}{@{}c|cccccccccc|cc@{}}
\toprule
$taskID$       & $\mathcal{T}_0$              & $\mathcal{T}_1$              & $\mathcal{T}_2$              & $\mathcal{T}_3$              & $\mathcal{T}_4$               & $\mathcal{T}_5$              & $\mathcal{T}_6$              & $\mathcal{T}_7$              & $\mathcal{T}_8$              & $\mathcal{T}_9$              &                              &                              \\ \cmidrule(r){1-11}
$runID$        & \cellcolor[HTML]{F2F2F2}0    & \cellcolor[HTML]{C9C9C9}1    & \cellcolor[HTML]{DBDBDB}2    & \cellcolor[HTML]{EDEDED}3    & \cellcolor[HTML]{F4B084}4     & \cellcolor[HTML]{F8CBAD}5    & \cellcolor[HTML]{FCE4D6}6    & \cellcolor[HTML]{8EA9DB}7    & \cellcolor[HTML]{B4C6E7}8    & \cellcolor[HTML]{D9E1F2}9    & \multirow{-2}{*}{$Acc_a$}    & \multirow{-2}{*}{$Acc_w$}    \\ \midrule
$\mathcal{P}_0$ & \cellcolor[HTML]{F2F2F2}93.3 & \cellcolor[HTML]{C9C9C9}36.7 & \cellcolor[HTML]{DBDBDB}61.7 & \cellcolor[HTML]{EDEDED}57.7 & \cellcolor[HTML]{F4B084}100.0 & \cellcolor[HTML]{F8CBAD}26.5 & \cellcolor[HTML]{FCE4D6}35.7 & \cellcolor[HTML]{8EA9DB}58.1 & \cellcolor[HTML]{B4C6E7}54.1 & \cellcolor[HTML]{D9E1F2}50.6 & \cellcolor[HTML]{D9E1F2}76.0 & \cellcolor[HTML]{D9E1F2}60.9 \\
$\mathcal{P}_1$ & \cellcolor[HTML]{D9E1F2}91.9 & \cellcolor[HTML]{F2F2F2}42.4 & \cellcolor[HTML]{C9C9C9}62.5 & \cellcolor[HTML]{DBDBDB}64.6 & \cellcolor[HTML]{EDEDED}99.3  & \cellcolor[HTML]{F4B084}24.5 & \cellcolor[HTML]{F8CBAD}60.3 & \cellcolor[HTML]{FCE4D6}56.2 & \cellcolor[HTML]{8EA9DB}62.2 & \cellcolor[HTML]{B4C6E7}61.8 & \cellcolor[HTML]{B4C6E7}74.0 & \cellcolor[HTML]{B4C6E7}61.7 \\
$\mathcal{P}_2$ & \cellcolor[HTML]{B4C6E7}92.6 & \cellcolor[HTML]{D9E1F2}42.7 & \cellcolor[HTML]{F2F2F2}60.6 & \cellcolor[HTML]{C9C9C9}66.1 & \cellcolor[HTML]{DBDBDB}100.0 & \cellcolor[HTML]{EDEDED}25.8 & \cellcolor[HTML]{F4B084}55.4 & \cellcolor[HTML]{F8CBAD}69.3 & \cellcolor[HTML]{FCE4D6}71.1 & \cellcolor[HTML]{8EA9DB}54.5 & \cellcolor[HTML]{8EA9DB}73.6 & \cellcolor[HTML]{8EA9DB}61.3 \\
$\mathcal{P}_3$ & \cellcolor[HTML]{8EA9DB}93.3 & \cellcolor[HTML]{B4C6E7}39.2 & \cellcolor[HTML]{D9E1F2}68.4 & \cellcolor[HTML]{F2F2F2}60.2 & \cellcolor[HTML]{C9C9C9}100.0 & \cellcolor[HTML]{DBDBDB}34.0 & \cellcolor[HTML]{EDEDED}57.3 & \cellcolor[HTML]{F4B084}71.4 & \cellcolor[HTML]{F8CBAD}85.2 & \cellcolor[HTML]{FCE4D6}58.3 & \cellcolor[HTML]{FCE4D6}74.9 & \cellcolor[HTML]{FCE4D6}63.4 \\
$\mathcal{P}_4$ & \cellcolor[HTML]{FCE4D6}89.6 & \cellcolor[HTML]{8EA9DB}34.4 & \cellcolor[HTML]{B4C6E7}64.0 & \cellcolor[HTML]{D9E1F2}59.9 & \cellcolor[HTML]{F2F2F2}100.0 & \cellcolor[HTML]{C9C9C9}35.1 & \cellcolor[HTML]{DBDBDB}65.8 & \cellcolor[HTML]{EDEDED}71.0 & \cellcolor[HTML]{F4B084}77.8 & \cellcolor[HTML]{F8CBAD}72.5 & \cellcolor[HTML]{F8CBAD}75.6 & \cellcolor[HTML]{F8CBAD}55.8 \\
$\mathcal{P}_5$ & \cellcolor[HTML]{F8CBAD}91.1 & \cellcolor[HTML]{FCE4D6}60.9 & \cellcolor[HTML]{8EA9DB}56.9 & \cellcolor[HTML]{B4C6E7}64.2 & \cellcolor[HTML]{D9E1F2}99.3  & \cellcolor[HTML]{F2F2F2}45.7 & \cellcolor[HTML]{C9C9C9}77.9 & \cellcolor[HTML]{DBDBDB}77.1 & \cellcolor[HTML]{EDEDED}67.4 & \cellcolor[HTML]{F4B084}79.9 & \cellcolor[HTML]{F4B084}75.1 & \cellcolor[HTML]{F4B084}58.3 \\
$\mathcal{P}_6$ & \cellcolor[HTML]{F4B084}94.1 & \cellcolor[HTML]{F8CBAD}66.9 & \cellcolor[HTML]{FCE4D6}72.1 & \cellcolor[HTML]{8EA9DB}55.5 & \cellcolor[HTML]{B4C6E7}100.0 & \cellcolor[HTML]{D9E1F2}53.0 & \cellcolor[HTML]{F2F2F2}77.2 & \cellcolor[HTML]{C9C9C9}74.3 & \cellcolor[HTML]{DBDBDB}84.4 & \cellcolor[HTML]{EDEDED}74.0 & \cellcolor[HTML]{EDEDED}74.5 & \cellcolor[HTML]{EDEDED}59.2 \\
$\mathcal{P}_7$ & \cellcolor[HTML]{EDEDED}94.8 & \cellcolor[HTML]{F4B084}73.6 & \cellcolor[HTML]{F8CBAD}85.4 & \cellcolor[HTML]{FCE4D6}53.6 & \cellcolor[HTML]{8EA9DB}98.6  & \cellcolor[HTML]{B4C6E7}61.6 & \cellcolor[HTML]{D9E1F2}80.4 & \cellcolor[HTML]{F2F2F2}81.5 & \cellcolor[HTML]{C9C9C9}92.6 & \cellcolor[HTML]{DBDBDB}86.5 & \cellcolor[HTML]{DBDBDB}76.0 & \cellcolor[HTML]{DBDBDB}62.0 \\
$\mathcal{P}_8$ & \cellcolor[HTML]{DBDBDB}94.8 & \cellcolor[HTML]{EDEDED}74.6 & \cellcolor[HTML]{F4B084}85.2 & \cellcolor[HTML]{F8CBAD}85.0 & \cellcolor[HTML]{FCE4D6}98.6  & \cellcolor[HTML]{8EA9DB}69.1 & \cellcolor[HTML]{B4C6E7}79.3 & \cellcolor[HTML]{D9E1F2}75.2 & \cellcolor[HTML]{F2F2F2}87.4 & \cellcolor[HTML]{C9C9C9}92.9 & \cellcolor[HTML]{C9C9C9}74.2 & \cellcolor[HTML]{C9C9C9}54.5 \\
$\mathcal{P}_9$ & \cellcolor[HTML]{C9C9C9}91.1 & \cellcolor[HTML]{DBDBDB}76.9 & \cellcolor[HTML]{EDEDED}91.3 & \cellcolor[HTML]{F4B084}89.4 & \cellcolor[HTML]{F8CBAD}97.9  & \cellcolor[HTML]{FCE4D6}65.4 & \cellcolor[HTML]{8EA9DB}77.6 & \cellcolor[HTML]{B4C6E7}83.9 & \cellcolor[HTML]{D9E1F2}93.3 & \cellcolor[HTML]{F2F2F2}85.8 & \cellcolor[HTML]{F2F2F2}76.1 & \cellcolor[HTML]{F2F2F2}58.9 \\ \midrule
$\mu$          & 92.7                         & 54.8                         & 70.8                         & 65.6                         & 99.4                          & 44.1                         & 66.7                         & 71.8                         & 77.6                         & 71.7                         & 75.0                         & 59.6                         \\
$\sigma$       & 1.73                         & 17.34                        & 12.22                        & 12.09                        & 0.77                          & 17.25                        & 14.61                        & 8.97                         & 13.36                        & 14.75                        & 0.91                         & 2.83                         \\ \bottomrule
\end{tabular}}
\caption{A case study of CML on the FewRel dataset.}
\label{CML-roll}
\end{table*}